\documentclass[letterpaper, 10 pt, journal]{ieeetran}  

\usepackage{graphicx}
\usepackage{amsmath}
\usepackage{amssymb}
\usepackage{blkarray}
\usepackage[noend]{algpseudocode}
\usepackage[ruled,vlined,linesnumbered]{algorithm2e}
\usepackage{setspace}
\usepackage{cite}
\usepackage[nolist]{acronym}
\usepackage{cite}
\usepackage{xcolor}
\usepackage{hyperref}
\usepackage{float}
\usepackage[thinc]{esdiff}
\def\BState{\State\hskip-\ALG@thistlm}

\SetKwInput{KwInput}{Input}

\DeclareMathOperator*{\argmax}{argmax}

\title{\LARGE \bf	Structurally aware 3D gas distribution mapping using belief propagation: a real-time algorithm for robotic deployment}

\author{Callum Rhodes$^{1}$, Cunjia Liu$^{1}$ and Wen-Hua Chen$^{1}$ 
	\thanks{This work was supported in part by EPSRC DTP under the project No. 2126619}
	\thanks{$^{1}$The authors are with the Department of Aeronautical and Automotive Engineering, Loughborough University, LE11 3TU, UK.		{\tt\small \{C.Rhodes, C.Liu5, W.Chen\}@lboro.ac.uk}}%
}

\begin{document}

\maketitle
\thispagestyle{empty}
\pagestyle{empty}

\begin{acronym}
\acro{CBRN}{chemical, biological, radiological and nuclear}
\acro{GDM}{gas distribution mapping}
\acro{GMRF}{Gaussian Markov Random Field}
\acro{GaBP}{Gaussian belief propagation}
\acro{UAV}{unmanned aerial vehicle}
\acro{UGV}{unmanned ground vehicle}
\acro{MRO}{mobile robot olfaction}
\acro{GP}{Gaussian process}
\acro{CFD}{computational fluid dynamics}
\acro{DSTL}{Defence Science and Technology Laboratory}
\acro{RMSE}{root mean square error}
\acro{SLAM}{simultaneous localisation and mapping}
\acro{IPP}{informative path planning}
\acro{STE}{source term estimation}
\acro{RGaBP}{residual belief propagation}
\acro{MAP}{maximum a posteriori}
\acro{PID}{photoionisation detector}
\acro{IPU}{intelligence processing unit}
\end{acronym}

\begin{abstract}
This paper proposes a new 3D gas distribution mapping technique based on the local message passing of Gaussian belief propagation that is capable of resolving in real time, concentration estimates in 3D space whilst accounting for the obstacle information within the scenario, the first of its kind in the literature. The gas mapping problem is formulated as a 3D factor graph of Gaussian potentials, the connections of which are conditioned on local occupancy values. The Gaussian belief propagation framework is introduced as the solver and a new hybrid message scheduler is introduced to increase the rate of convergence. The factor graph problem is then redesigned as a dynamically expanding inference task, coupling the information of consecutive gas measurements with local spatial structure obtained by the robot. The proposed algorithm is compared to the state of the art methods in 2D and 3D simulations and is found to resolve distribution maps orders of magnitude quicker than typical direct solvers. The proposed framework is then deployed for the first time onboard a ground robot in a 3D mapping and exploration task. The system is shown to be able to resolve multiple sensor inputs and output high resolution 3D gas distribution maps in a GPS denied cluttered scenario in real time. This online inference of complicated plume structures provides a new layer of contextual information over its 2D counterparts and enables autonomous systems to take advantage of real time estimates to inform potential next best sampling locations. \\
\end{abstract}

\def\abstractname{Note to Practitioners}

\begin{abstract}
The motivation of this work arises from the need to develop gas distribution mapping tools for mobile robots that can provide real-time situational awareness. The output distribution maps can be used to inform human first responders as to what areas of the environment contain a hazard, but looking towards autonomous robots, they can also be used by the robot itself to inform where should be measured next to gather more information about the hazard. When performing these mapping tasks in unknown indoor environments, it is very important that the sensing robot can build up knowledge of its physical surroundings together with the how the obstacles in the environment effect the gas distribution, in full 3D. The Gaussian belief propagation algorithm allows us to achieve all of this in real-time onboard the sensing robot, something that is yet to be achieved in the literature. \\
\end{abstract}


\section{Introduction}
\IEEEPARstart{A}{utonomous} robots have seen a dramatic increase in research in recent years due to their ability to perceive environments without human intervention. In this regard, the use of mobile robotics within \ac{CBRN} applications is of great modern interest. Recent disaster events, such as the Fukushima disaster in 2011, and the ever present threat of chemical warfare have dictated the need for quickly deployable systems that can identify the characteristics of a gas dispersion event whilst operating in potentially unknown environments. In such scenarios, key media inferred by mobile systems can inform first responders whilst simultaneously providing safe data collection. 	
 
 Gas distribution mapping answers the proposed question of \textit{How is the gas distributed across an area?} Whilst in certain scenarios it may only be of interest to determine if a gas is present or where the source of the release is, the gas distribution across the whole environment can provide richer information. Contextually, distribution maps be can used to identify areas of high concentration (as well as low concentration) in order to dictate exclusions zones, the boundary of the plume and likely locations of a leak. This information is a key first step in hazard response. Furthermore, inferring distribution maps in real time as data are collected can help to determine the next best sampling location for a sensor in both the autonomous and teleoperated cases. 
	
	A combination of small scale fluctuations in the dispersal pattern, a limited sensing area and transient measurements associated with chemical sensors makes gas mapping an exceptionally difficult task compared to the similar \ac{CBRN} related field of radiation mapping, which can rely on a smooth inverse-squared relationship to make predictions \cite{Towler2012}. Equally, more complicated model based methods that utilise \ac{CFD} (e.g. \cite{Branford2011}) are capable of producing high fidelity distributions that can account for complex geometry. However, these methods rely on reliable prior meteorological information and can take in the order of days to converge to an accurate solution. This makes them unsuitable for both first response and mobile sensor data collection which both require solutions in real time. Based on these insights, there has been an increasing trend towards investigating data driven methods for mapping a gas distribution. These methods make no strong assumptions on the gas transport model and instead rely on online fitting of a function to dense and diverse data.

\subsection{Related works}

In the literature, there are three popular data-driven \ac{GDM} methods that have been designed for use specifically with mobile robotic sensors, which are Kernel DM (and its variants) \cite{Lilienthal2003, Lilienthal2009, reggente2009, Plagemann2011}, \ac{GMRF} \cite{G.Monroy2016, Gongora2020} and \ac{GP} \cite{Stachniss2009, Hutchinson2019}.

Kernel DM, first proposed by Lilienthal in 2003 \cite{Lilienthal2003}, is one of the first attempts to tackle the problem of concentration mapping for mobile robotics. This is achieved by discretising a spatial map into a grid array with each cell in the grid storing the mean concentration predicted by the Kernel algorithm. Each measurement then applies a Gaussian shaped contribution to the surrounding cells based on the distance away from the measurement. By formulating the problem in this manner, concentrations in the environment are only dependent on the measurement themselves and not their surrounding cells. The work has been extended in \cite{Lilienthal2009}, where Kernel DM+V provides a distribution of the confidence and the variance, where variance relates to the local variability of concentrations (source term predictor) and confidence relates to the uncertainty of the mean distribution. In \cite{reggente2009}, Kernel DM+V/W is developed. In this work, the Gaussian kernel, is transformed into a bivariate Gaussian along the axis of a measured wind direction, thus enabling the inclusion of anemometry sensor measurements. A further addition to the kernel method is shown in \cite{Plagemann2011}, where TD kernel DM+V/W allows for the time decay of measurements as they become historically uncertain. These works outline the core of the kernel DM algorithm, however further works have investigated multi-compound discrimination \cite{Bennetts2014}, a 3D kernel \cite{Luo2015} and integrating obstacle effects \cite{Visvanathan2020a}. The obstacle inclusion works by scaling the contribution from a measurement to a cell by the Euclidean distance transform over the map from the measurement to the cell. This must be calculated for each measurement and also each cell within its radius and therefore is computationally inefficient for high resolution maps or 3D mapping tasks. Whilst Kernel DM is a lightweight local \ac{GDM} algorithm, the method does not implicitly include the effects that obstacles have on the plume in a statistical manner due to the fundamental lack of dependency between predictions and the spatial map.

The \ac{GP} technique as a general regression model, has also been applied to \ac{GDM} in early work by Stachniss et. al \cite{Stachniss2009} and more recently by Hutchinson et. al \cite{Hutchinson2019}. In the earlier example, a Gaussian Process mixture model is employed wherein a mixture of 2 Gaussian processes are simultaneously learnt with one for the smooth background concentration function and one to model the plume concentration. The Matern 3/2 covariance function is used for the shaping kernel. The mixture model is compared against the Kernel DM method and single \ac{GP} technique in an experimental trial and the performance of mixture model is shown to be the best at predicting the gas distribution. \ac{GP} models are capable of predicting continuous distributions over a domain, contrary to discrete methods, and therefore are not susceptible to discretisation errors. However, \ac{GP}s scale poorly with an increasing number of measurements (which is necessary for large scale mapping tasks) and are not currently capable of accounting for obstacle integration with the predicted map. 

Most recently, the \ac{GMRF} algorithm for \ac{GDM} has been proposed by Monroy et. al \cite{G.Monroy2016} that discretises the environment into cells and then forms a graphical model that describes the relationship between adjacent cells and measurements. In fact, it is shown in that a \ac{GMRF} is equivalent to a \ac{GP} in \cite{Choi2012}. Contrary to the \ac{GP}, due to modelling the dependency between a cell and its neighbours, the \ac{GMRF} model probabilistically accounts for the effects of obstacles on the concentration at a given cell whilst also accounting for the increasing uncertainty of measurements over time (similarly to TD Kernel DM). The underlying graphical model used is known as a factor graph and is a common representation of inference problems in mobile robotics \cite{Dellaert2017}. Due to the Gaussian assumptions placed upon the factor graph, the problem reduces to a \ac{GMRF} that can be solved by the least squares method. In \cite{Gongora2020}, the algorithm is extended to also model the wind field via measured anemometer readings and resolve the effects that a wind field has on a cells gas concentration (bringing parity with TD Kernel DM+V/W). The major downside of G-\ac{GMRF} stated in \cite{G.Monroy2016}, is that the computational complexity is $\mathcal{O}(n^{1.5})$ and the entire map must be resolved per iteration. This limits the size of map that a mobile robot can calculate and also requires that the map is fully resolved each time an update to the graph is added. 

To overcome this issue, Rhodes et. al \cite{Rhodes2022b} advocate the use of the \ac{GaBP} solver on the \ac{GDM} factor graph problem. Belief propagation is a message passing technique that can be applied to loopy graph structures \cite{Murphy2013a} (such as those factor graphs in \ac{GDM}) and has seen increasing prominence in mobile robotics in recent years, especially in the SLAM domain \cite{Ranganathan2007, Davison2019, Ortiz2020}. The G-\ac{GaBP} algorithm proposed makes use of the wildfire message schedule \cite{Ranganathan2007} to iteratively solve the factor graph via local message passing propagating around sensor measurements. This shift away from global solvers (e.g., the direct method used in \cite{G.Monroy2016}) towards local inference allows for orders of magnitude quicker solutions to sequential mapping, and can also be used to acquire anytime estimates of the distribution maps (as opposed to waiting for a batch solution). Whilst the initial work on G-\ac{GaBP} proposes the basic application of belief propagation to the \ac{GDM} domain, simulation work is only carried out on a 2D scenario and inference is still performed on a globally defined map. 

Therefore, we extend our work by addressing the problem of performing inference on a 3D gas distribution map, applying several efficiency upgrades, and also showing the real-world applicability of the algorithm to an online mapping robotic platform. We now list the specific contributions of the work displayed in this paper.

\subsection{Contributions}
This paper consists of four main contributions to the state of the art. Firstly, we formulate the \ac{GDM} problem as a 3D network of connected states and measurements from which we can perform probabilistic inference, increasing functionality from the 2D state of the art.

We then propose two major improvements to the \ac{GaBP} solver proposed in \cite{Rhodes2022b} to increase its efficiency when applied to mapping with multiple point sampling sensors. The first improvement concerns the schedule for sending messages in the belief propagation algorithm. A hybrid solver is proposed that utilises the properties of both the wildfire and residual belief propagation algorithms to increase the global convergence rate. The second improvement reformulates the inference problem from a static a-priori graph into a dynamic information theoretic graph structure that grows relative to the information inferred.

Our final contribution is the application of the proposed solver into a robotic framework that is capable of performing real-time inference in completely a-prior unknown environments. By doing so, we show the first mobile robot capable of performing 3D gas distribution mapping in conjunction with a dynamic occupancy map (to inform factor graph construction), whilst also performing real time \ac{SLAM}. All the above functions are shown being performed entirely onboard the robotic platform with real-time inference updates available to a potential operator or informative path planner. We also make the deployed algorithm available as on open source ROS package for others to implement on their systems.

The remainder of the paper is as follows. Section \ref{sec:problemform} formulates the \ac{GDM} problem as a 3D factor graph and shows how it can be simplified to a \ac{GMRF} under Gaussian assumptions. We then introduce the governing equations behind the \ac{GaBP} framework in Section \ref{sec:beliefpropa} and describe how the \ac{GMRF} problem can be solved using local message passing. Next, the proposed 3D G-\ac{GaBP} algorithm is derived in Section \ref{sec:3dsolver} detailing the aforementioned developments to the propagation framework. The proposed algorithm is then tested in several simulations in Section \ref{sec:sim} for quantitative analysis. Finally, the experimental evaluation is described in Section \ref{sec:experiment}, the proposed algorithm is deployed onboard a robotic platform in a real-world 3D mapping task.

\section{Problem Formulation} \label{sec:problemform}

Let $\mathcal{X} = \{ x_1, \ldots, x_n \}$ be a finite set of random variables that we wish to infer with each variable $x_{i}$ representing a distribution of the concentration at the $i$-th cell or voxel in the domain $\mathcal{D} \subset \mathbb{R}^{3}$. We assume that the sensing robot takes a set of measurements at different locations within the domain. Denoting $z_{k,i}$ the noisy sensor measurement taken at time $k$ associated with $x_i$, the set of measurements corresponding to $x_i$ collected at different timestamps is denoted as $\mathcal{Z}_{i}$. With a slight abuse of notation, we use the subscript $k$ as the index of the measurement within the set $\mathcal{Z}_{i}$. The total measurement set collected so far is expressed as $\mathcal{Z} = \{\mathcal{Z}_{i} \mid  \mathcal{Z}_{i} \neq \emptyset,\, i = 1,\ldots, n \}$. We aim to resolve the \ac{MAP} estimate of the variables $\mathbf{x} = \begin{bmatrix} x_1 \cdots x_n \end{bmatrix}^T$ from the joint probability density function $p(\mathbf{x} | \mathcal{Z})$ to give the most likely cell concentrations across the gas mapping domain.

\subsection{Factor graphs} \label{sec:factorGraphs}

To model the dependencies between these random variables and the measurements recorded, the \ac{GDM} problem can be formulated as an inference problem over a factor graph $G = (\mathcal{X}, \mathcal{F}, \boldsymbol{\varepsilon})$, which is an undirected bi-partite graph containing variable nodes $\mathcal{X}$, factor nodes $\mathcal{F} = \{f_1,f_2,\ldots,f_S\}$ and their edges $\boldsymbol{\varepsilon}$ encoding the connectivity according to the variables involved in each factor. Each factor $f_s(\cdot)$ describes the dependencies between a local subset of variable nodes $\mathcal{X}_s \subset \mathcal{X}$, which are directly connected to the factor. Considering all nodes in the graph, the joint probability distribution is expressed as the product of all factors in the graph \cite{Dellaert2017}:
\begin{align} \label{eqn:prod}
    p(\mathbf{x} | \mathcal{Z}) \propto \prod_{s=1}^{S} f_s(\mathcal{X}_s).
\end{align}

Factors in the general sense can take any form, however, it is typical in mobile robotics to represent factors as Gaussian distributions, since it is a common assumption that sensor measurements deviate from the ground truth value in a Gaussian manner and that the gas concentration values also affect their surroundings spatially in a Gaussian manner, as in other \ac{GDM} algorithms (e.g. Kernel DM and \ac{GP}). When applying the Gaussian assumption to all factors in the factor graph, it can now be represented as a \ac{GMRF}. From this model, the problem of retrieving the \ac{MAP} solution of vector $\mathbf{x}$ is the same as finding the marginal means of the associated Gaussian joint probability distribution being $p(\mathbf{x}|\mathcal{Z}) = \mathcal{N}(\mathbf{\Lambda}^{-1}\mathbf{g}, \mathbf{\Lambda}^{-1})$, i.e.,
\begin{align}
    p(\mathbf{x}|\mathcal{Z}) \propto \exp(\mathbf{x}^T\mathbf{\Lambda}\mathbf{x}/2-\mathbf{g}^T\mathbf{x}),
\end{align}
where $\mathbf{\Lambda}$ is the Information matrix of the \ac{GMRF} model and $\mathbf{g}$ is the observation vector \cite{Dellaert2017}. The construction of such a model for \ac{GDM} is derived in section \ref{sec:GDMfactors}. 


By factorising the pairwise cliques defined by the factors in the graph between variable $x_i$ and $x_j$, Eq. \eqref{eqn:prod} can be rewritten in \ac{GMRF} form as a function of edge potentials, $\psi_{ij} \triangleq \text{exp}(-x_i\mathbf{\Lambda}_{ij}x_j) $, and self potentials, $\phi_i \triangleq \text{exp}(\mathbf{g}_ix_i - \mathbf{\Lambda}_{ii}x_i^2 / 2)$, such that
\begin{align} \label{eqn:prod_GMRF}
		p(\mathbf{x}) \propto \prod_{i=1}^{n}\phi_i(x_i) \prod_{\{i,j\}}\psi_{ij}(x_i,x_j).
\end{align} 

\subsection{Factor graph construction for 3D-GDM} \label{sec:GDMfactors}

In our previous work \cite{Rhodes2022b}, we make use of the typical 2D graph construction of Monroy et al. \cite{G.Monroy2016}. We now extend the factor graph formulation to account for 3D gas distribution maps. For the basic \ac{GDM} problem, we take three types of Gaussian factors that represent the correlations between concentration variables and some observed quantity, including the relationship between neighbouring discrete cells in $\mathbb{R}^3$ space. These are observation factors $f_{o}$, regularisation factors $f_r$ and default factors $f_d$.

\begin{figure*}[t]
    \centering
    \includegraphics[width=0.75\textwidth]{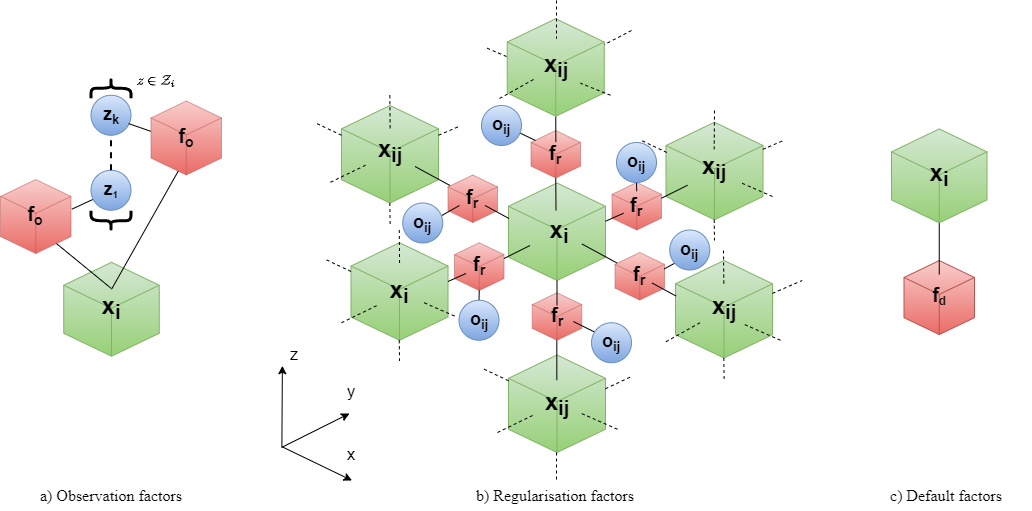}
    \caption{Factor graph cliques for each type of factor in the 3D-\ac{GDM} problem.}
    \label{fig:gdmFactors}
\end{figure*}

The observation factor defines the potential between a noisy sensor measurement $z_{k,i}$ and the modelled concentration value $x_i$. Since the measurements are observed variables and not part of the inference itself, these potentials contribute to a nodes self potential $\phi_i$ and are often referred to as unary factors. The sensor measurement is modelled as $z_{k,i} = x_i + \omega_k + \zeta_k$ with two additive Gaussian noises, of which $\omega_k \sim \mathcal{N}(0, \sigma^2_s)$ arises from the sensor noise and $\zeta_k \sim \mathcal{N}(0, \sigma^2_\zeta \Delta t)$ is from the time dependent corruption that models the increase in uncertainty over time. $\sigma^2_\zeta$ should be set relative to the volatility of the underlying gas distribution (i.e. higher for more volatile distributions and lower for stable plumes). The exponential energy function $E$ associated with the Gaussian factor $f_o$ for a single $z_{k,i}$ is therefore given as $E_o(z_{k,i}) = 0.5(x_i - z_{k,i})^2/(\sigma_s^2 + \sigma_\zeta^2 \Delta t_{z_{k,i}})$,
where $\Delta t_{z_{k,i}}$ is the time difference between the current time step and the time the measurement $z_{k,i}$ was recorded. The factor for each node $x_{i}$ with a non-empty set of measurements $\mathcal{Z}_i$ can be expressed as 
\begin{equation}\label{eqn:factor_o}
    f_{o}(x_i) \propto \exp{\left( \sum_{k=1}^{|\mathcal{Z}_{i}|} - \frac{1}{2} \alpha_{k,i} x_i^2 + z_{k,i} \alpha_{k,i} x_{i} \right)},
 \end{equation}
where $|\mathcal{Z}_{i}|$ refers to the cardinality of the measurement set associated with the $i$-th node and $\alpha_{k,i} = 1/(\sigma_s^2 + \sigma_\zeta^2 \Delta t_{z_{k,i}})$.


The regularisation factor $f_r$ defines the potential between the connected nodes in the graph and relates to the influence that the gas concentration at voxel $i$ has on each its neighbours $j \in \mathcal{N}_i$, given as $l_{i,j} = x_i - x_j$. In the 3D  case, the regularisation factor is also defined between voxels that are above and below $x_i$ (as shown in Fig. \ref{fig:gdmFactors}) and thus $|\mathcal{N}_i| \leq 6$. The associated Gaussian for regularisation is modelled by $l_{i,j} \sim \mathcal{N}(0,\sigma^2_r)$. By tuning $\sigma^2_r$ relative the $\sigma^2_o$, we scale how much a measurement will affect its surrounding nodes. Regularisation factors also account for the effect that an obstacle between nodes has on the distribution by conditioning the factor graph based on an occupancy map. Let $o_{ij} \in \{0,1\}$ refer to the binary occupancy between node $i$ and its neighbour $j$. When $o_{ij}=1$, there is an obstacle blocking the correlation of the two cells' concentrations. If $o_{ij}=0$, free space exists between the nodes, and therefore the regularisation factor acts uninterrupted. This gives the governing energy equation for regularisation of a pair of nodes ($l_{i,j}$) that are conditioned on the occupancy map as $E_r(l_{i,j}) = 0.5(x_i - x_j)^2/(\sigma_r^2 / (1 - o_{ij})^{2})$. Thus, the regularisation factor associated with node $i$ can be written as 
\begin{equation} \label{eqn:factor_r}
    f_{r}(x_i) \propto \exp{ \left(  \sum_{j \in \mathcal{N}_i} - \frac{1}{2} \beta_{i,j} x_{i}^2 + x_j \beta_{i,j} x_i - \frac{1}{2} \beta_{i,j} x_j^2  \  \right) },
\end{equation}
where $\beta_{i,j} = 1/(\sigma_r^2 / (1 - o_{ij})^{2})$. Note that the regularisation factor contribute to both the edge potentials and the self-potential of the related nodes. 

Finally, the default factor $f_d$ is defined which lightly anchors the concentrations at cells to a nominal level ($z_{0,i}$), given by the Gaussian $x_i \sim \mathcal{N}(z_{0,i},\sigma^2_d)$. This is required to enforce the cell's concentration close to zero (or any other arbitrary background concentration value) in the absence of any information in its surrounding area. If this is not set, then sensor measurements can propagate infinitely far into the factor graph causing high predictions in unsampled regions. This factor will also improve the numerical stability for isolated local cells. By tuning this factor relative to $f_o$, we can scale how much a sensor measurement overrides the zero prior value. This factor is similar to $f_o$ in that it contributes to a cells self potential. The energy related to the default factor for node $x_i$ is given as $E_d(z_{0,i}) = 0.5(x_i - z_0)^2/\sigma_d^2$, and thus the default factor for all nodes can be expressed as
\begin{equation} \label{eqn:factor_d}
    f_d(\mathbf{x}) \propto \exp{ \left( \sum_{i=1}^{n} - \frac{1}{2\sigma_d^2}x_i^2 + \frac{z_{0,i}}{\sigma_d^2} x_i \right)}.
\end{equation}

Based on the established definitions of the individual factors, the information matrix $\mathbf{\Lambda}$ and the observation vector $\mathbf{g}$ of the \ac{GDM} problem can be constructed. Collecting the first term in \eqref{eqn:factor_o}-\eqref{eqn:factor_d}, it can be derived that the diagonal entries of the information matrix $\mathbf{\Lambda}$ follows
\begin{equation} \label{eqn:info_ii}
    \mathbf{\Lambda}_{ii} = \frac{1}{\sigma^2_d} + \sum_{k = 1}^{|\mathcal{Z}_i|} \frac{1}{\sigma_s^2 + \sigma_\zeta^2 \Delta t_{z_{k,i}}} + \sum_{j \in \mathcal{N}_i} \frac{1}{\sigma_r^2 / (1 - o_{ij})},
\end{equation}
whereas the second term in \eqref{eqn:factor_o} and \eqref{eqn:factor_d} contributes to the $i$-th entry of $\mathbf{g}$ such that
\begin{equation} \label{eqn:g_i}
    \mathbf{g}_i =  \frac{z_{0,i}}{\sigma^2_d} + \sum_{k = 1}^{|\mathcal{Z}_i|} \frac{z_k}{\sigma_s^2 + \sigma_\zeta^2 \Delta t_{z_{k,i}}}. 
\end{equation}
The off-diagonal entries of the information matrix $\mathbf{\Lambda}_{ij}$ depend on the connectivity of the graph. The non-zero entries are associated to the edges of the graph and can be formed as
\begin{equation} 
    \mathbf{\Lambda}_{ij} = \mathbf{\Lambda}_{ji} = \frac{1}{2 \sigma_r^2 / (1 - o_{ij})}, \quad \{i,j\} \in \boldsymbol{\varepsilon}.
\end{equation}

As \eqref{eqn:info_ii} - \eqref{eqn:g_i} fully define the 3D-\ac{GDM} \ac{GMRF} model, an appropriate solver is then needed to provide the solution to the inference problem. The marginal inference problem for each cell focuses on $p(x_i) = \mathcal{N}(\mu_i, P_i^{-1})$, where $\mu_i=\{\mathbf{\Lambda}^{-1}\mathbf{g}\}_i$ is the mean and $P_i$ is the inverse of the variance (i.e., precision) of node $i$, given by $P_i^{-1}=\{\mathbf{\Lambda}^{-1}\}_{ii}$. 

It should be noted that with the increased size of $\mathcal{N}_i$ to account for the 3D problem, inference will be performed on an information matrix significantly less sparse than the 2D case. Furthermore, unless a larger discretisation size is chosen, $|\mathbf{\mathcal{X}}|$ will also significantly increase. Therefore, the combination of these two characteristics means that solutions with direct solvers very quickly become infeasible. For large scale environmental monitoring, more efficient solvers must be implemented in order to retain online inference despite the scale of the problem. 

\textcolor{black}{In this implementation we model gas distribution characteristics using only concentration measurements, despite both \cite{reggente2009} and \cite{Gongora2020} showing that including anemometer measurements can provide more accurate maps. Our main focus experimentally is on indoor locations where the wind field is not significant enough to dominate gas transport, so we instead optimise the solver for \ac{GDM}, leaving the extension of modelling the wind field as future work.} 



\section{Gaussian belief propagation for GDM} \label{sec:beliefpropa}

In \cite{G.Monroy2016}, after establishing the \ac{GMRF} model, the solution of the inference is then calculated as the convenient least squares problem which in the original work is solved using the Cholesky decomposition to solve the Gauss-Newton method equations. There are many different solutions to inference on a \ac{GMRF}, as discussed in \cite{Dellaert2017}, and by utilising a solver that suits the application of a mobile robotic sampling agent, real time estimates of complex distributions can be achieved. 

\begin{figure}[t]
    \centering
    \includegraphics[width=0.45\columnwidth]{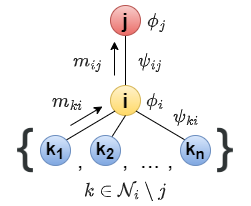}
    \caption{Local message passing from node $i$ to its neighbour node $j$.}
    \label{fig:nodeBP}
\end{figure}

Belief propagation is a general tool for inference on a factor graph and functions by passing iteratively updating belief messages between neighbourhoods in the factor graph. This message passing process is equivalent to Pearl's local messsage-passing algorithm \cite{Pearl1988}. Due to the Gaussian assumption placed upon the factor graph, a message passed between nodes is a Gaussian belief. The following description is based the procedure by Shental et. al \cite{Shental2008} and an excellent visual overview of belief propagation by Ortiz et. al can be found in \cite{Ortiz2021}. A message consists of real values that are passed between nodes and are governed by two rules: the `sum-product rule' and the `product rule'. Since we are dealing with a Markov random field of continuous variables, the sum-product rule becomes the integral-product rule \cite{Weiss2000}. Therefore, for a message sent from node $i$ to node $j$ the following applies:
\begin{align} \label{eqn:intprod}
	m_{ij}(x_j) \propto \int_{x_i}\psi_{ij}(x_i,x_j)\phi_i(x_i) \prod_{k\in \mathcal{N}_i\setminus j}m_{ki}(x_i)dx_i,
\end{align}
where the notation $\mathcal{N}_i\setminus j$ is the set of neighbours of $i$ not including node $j$. The marginals are computed as per the product rule:
\begin{align} \label{eqn:margprod}
	p(x_i) = \gamma \phi(x_i) \prod_{k\in \mathcal{N}_i}m_{ki}(x_i),
\end{align}
where $\gamma$ is a normalisation constant. Fig. \ref{fig:nodeBP} shows an example message passing between node $i$ and node $j$ in a small neighbourhood, with symbols corresponding to equations \eqref{eqn:intprod} and \eqref{eqn:margprod}. The self potentials established in section \ref{sec:factorGraphs} are of Gaussian form so that $\phi_{i} \propto \mathcal{N}(\mu_{ii}, , P^{-1}_{ii})$ and $m_{ki} \propto \mathcal{N}(\mu_{ki}, P^{-1}_{ki})$. Then based on the rules of the product of Gaussians shown in \cite{Shental2008}, the Gaussian associated with right hand side of equation \eqref{eqn:intprod} can be calculated as:
\begin{align} \label{eqn:messageUpdate}
    	P_{i \setminus j} &= \overbrace{P_{ii}}^{\phi_i} + \sum_{k\in \mathcal{N}_i\setminus j} \overbrace{P_{ki}}^{m_{ki}(x_i)} \\
    	\mu_{i\setminus j} &= P^{-1}_{i \setminus j} \big( \overbrace{P_{ii}\mu_{ii}}^{\phi_i} + \sum_{k\in \mathcal{N}_i\setminus j} \overbrace{P_{ki}\mu_{ki}}^{m_{ki}(x_i)} \big),
\end{align}
where $P_{ii} \triangleq \mathbf{\Lambda}_{ii}$ (a-priori precision) and $\mu_{ii} \triangleq \mathbf{g}_i / \mathbf{\Lambda}_{ii}$ (mean of the self potential). After resolving the remaining Gaussian integral of equation \eqref{eqn:intprod}, the message $m_{ij}(x_i)$ that is sent along an edge for any given pair of nodes in the network is given as
\begin{align} \label{eqn:messageSend}
    m_{ij}(x_i) \begin{cases}
        P_{ij} &= -\mathbf{\Lambda}_{ij}^2 P^{-1}_{i \setminus j} \\
        \mu_{ij} &= -P_{ij}^{-1} \mathbf{\Lambda}_{ij} \mu_{i \setminus j}.
    \end{cases}
\end{align}

Eq. \eqref{eqn:messageSend} shows how updated beliefs are sent from one node in the network to a neighbour, and then the next step is to repeatedly send messages between nodes under a specific scheduling until the messages reach a convergence threshold. The final step is to calculate the marginals of all $x_i \propto \mathcal{N}(\mu_i, P^{-1}_i) \in \mathcal{X}$ as per equation \eqref{eqn:margprod} to give:
\begin{align} 
    	P_{i} &= \overbrace{P_{ii}}^{\phi_i} + \sum_{k\in \mathcal{N}_i} \overbrace{P_{ki}}^{m_{ki}(x_i)} \\
    	\label{eqn:marginalmean}
    	\mu_{i} &= P^{-1}_{i} \big( \overbrace{P_{ii}\mu_{ii}}^{\phi_i} + \sum_{k\in \mathcal{N}_i} \overbrace{P_{ki}\mu_{ki}}^{m_{ki}(x_i)} \big).
\end{align}

This completes the \ac{GaBP} process and shows that since messages are passed locally with no specific ordering and that marginal mean and uncertainty values are calculated iteratively, the solver can flexibly resolve in any local region of the graph and recover the current solution vector at any time. This is a key motivation for using the \ac{GaBP} solver in the mobile robotic domain wherein information is usually acquired within a local region and can be without a fixed frequency.  

Comparing the estimation accuracy of such iterative solvers against their direct counterparts, the findings of Weiss et al. \cite{Weiss2000} show that the \ac{GaBP} solver will converge to the exact marginal means estimated by direct methods, even when the graph exhibits many loops (which is strongly the case in \ac{GDM} given the interconnected nature of the spatial distribution). However, in these very loopy graphs the variances will often be incorrect (always overconfident in \ac{GMRF}s), but as stated in \cite{Weiss2000}, for the marginals to converge to the exact values then the variances must be heuristically correct, indicating that despite their real values being incorrect, the relative values of the variances between nodes in the network must be correct. To further increase computational efficiency, the number of messages sent by \ac{GaBP} can be reduced from $\mathcal{O}(n^2)$ to $\mathcal{O}(n)$ per iteration round when broadcasting aggregated sum of potentials for each node, as per Algorithm 1 of \cite{Shental2008}.   

Since \ac{GaBP} is a general tool for solving factor graphs, the fact we are applying the solver to a \ac{GDM} problem or that it is a 3D-\ac{GDM} problem has no effect on its fundamental application. Therefore, equations \eqref{eqn:messageUpdate} - \eqref{eqn:marginalmean} can be directly applied to the 3D graph $\{\mathbf{\Lambda},\mathbf{g}\}$ derived in section \ref{sec:GDMfactors}.


\section{3D-GDM solver} \label{sec:3dsolver}

\begin{figure*}[ht]
    \centering
    \includegraphics[width=0.65\textwidth]{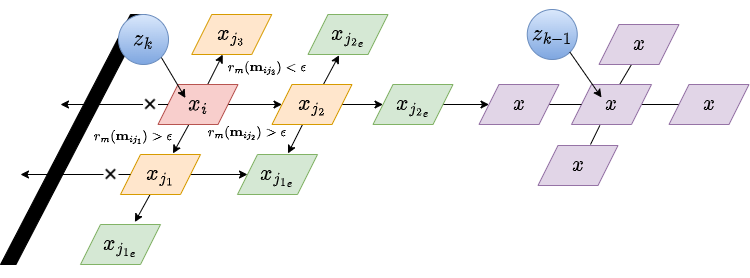}
    \caption{Information theoretic local factor graph growth around sequential measurements}
    \label{fig:graphGrowth}
\end{figure*}

Based on the formulation of the 3D factor graph and the principles of general Gaussian belief propagation, the proposed algorithm to achieve fast 3D inference will be outlined in this section. Key improvements over the current state-of-the-art include a hybrid message scheduler as well as a dynamically built factor graph based on an information theoretic criterion. With these changes in place, efficient local 3D-\ac{GDM} can be achieved.  

\subsection{Hybrid schedule with sporadic measurements}

In the introductory work on G-\ac{GaBP} \cite{Rhodes2022b}, several message scheduling algorithms are tested in a \ac{GDM} scenario. These are random, round-robin, residual \cite{Elidan2006} and wildfire \cite{Ranganathan2007} schedulers. It is found, for the specificity of environmental monitoring with a point sensor, that the wildfire algorithm has much better mean and variance convergence around the measurement region than the other schedules. It has also been found however that in unsampled regions away from the local vicinity, that the speed of uncertainty convergence is much slower when using the wildfire algorithm. In contrast, residual belief propagation shows good uncertainty convergence globally but with a poor rate of marginal mean convergence. Furthermore, when performing wildfire message propagation, messages are sent until a threshold information theoretic criteria is met, which leads to variable iteration times (depending on the amount of new information a particular new measurement contributes to the factor graph). Thus, there may be unused computation time between sensor readings where the inference has adequately converged in a local vicinity. This is especially relevant in mobile \ac{GDM} where concentration readings are often subject to a transient response \cite{Liu2012} and agents may need to stop and sample a location for a period of time \cite{Hutchinson2019a,Rhodes2022}. It would be useful in this downtime to continue resolving the factor graph in order to maximise the online performance of the mapping agent. Based on these insights, we propose a new hybrid schedule that leverages the rapid local convergence of the wildfire algorithm, combined with the global convergence of the residual belief propagation schedule. 

To achieve this, we enforce a mode switching that changes the message schedule to residual propagation when a wildfire iteration finishes. Since both wildfire and residual propagation track the same message residuals, extra computation is not required and the solver can simply switch between modes given its current convergence. The residual of a message is defined as the Bhattacharyya distance between the current message being sent and the previous message sent along the same edge:
\begin{align} \label{eqn:resid}
    r_m(m_{ij}) &= \|m_{ij}^{k} - m_{ij}^{k-1}\|_{B}  \\
        \begin{split}
                &= \frac{1}{4} \ln\bigg(\frac{1}{4}\bigg(\frac{P_{ij}^{k}}{P_{ij}^{k-1}} + \frac{P_{ij}^{k-1}}{P_{ij}^{k}} + 2 \bigg)\bigg) \\
                &+ \frac{1}{4}\bigg((P_{ij}^{k-1} + P_{ij}^{k})(\mu_{ij}^{k-1} - \mu_{ij}^{k})^2\bigg)
        \end{split}
\end{align}

Other information theoretic metrics can also be used and as discussed in \cite{Rhodes2022b}, but performance of the solver is more affected by the schedule rather than the specific metric (for the case of \ac{GDM}). Formally, the hybrid message scheduler is outlined in Alg. \ref{alg:sched}.

\begin{algorithm}
    \label{alg:sched}
    \SetAlgoLined
    \caption{Hybrid message schedule}
    \While{\text{Active}}{
        \tcp{Wildfire loop}
        \If{New $z_{k,i}$}{
        $Q \xleftarrow{} i$ \\ 
            \While{$Q \neq \emptyset$}{
                $t \gets Q(1)$ \\
                Update and send $m_{tj}$ according to Eq. \eqref{eqn:messageUpdate}-\eqref{eqn:messageSend} $\forall j \in \mathcal{N}_t$\\
                \If{$r_m(m_{tj}) > \epsilon$, \eqref{eqn:resid}}{
                    $Q \xleftarrow{+} j$
                }
                $Q \xleftarrow{-} t$ \\
            }
        }    
        \tcp{Residual propagation loop}
        $m_{tj} = \argmax_m r_m(\mathbf{m})$ \\ 
        Update and send $m_{tj}$ according to Eq. \eqref{eqn:messageUpdate}-\eqref{eqn:messageSend}
    } 
\end{algorithm}

Due to the nature of belief propagation wherein the marginal mean and precision values are calculated as part of the message passing process, the current global marginal means and variances can be extracted at any time during both the wildfire and residual loops. The efficiency gain of using a hybrid message schedule is tested in simulation in section \ref{sec:sim}. As stated in Introduction, this addition is a key contribution to efficient \ac{GDM} and is combined with our third contribution to the algorithm, to be described in the following subsection.

\subsection{Local graph building for local inference} \label{sec:local}

Thus far, our work on G-\ac{GaBP} has focused on how to utilise an efficient local solver in order to overcome the computational inefficiencies of the direct methods. Despite the shift to local inference, until now the factor graph upon which we perform inference is still in a global frame and considers the entire area to be searched. This is naturally opposed to the local nature of \ac{GaBP} and is something we wish to address.

In the works of \cite{G.Monroy2016, Gongora2020, Rhodes2020, Rhodes2022b}, the area is first discretised into cells and then the factor graph is built in its entirety upon this a-priori discretisation. This means that even with \ac{GaBP}, inference is being performed in areas far away from the sensor location and efforts are wasted in calculating these distant marginals which approach $\{\mu_i=0, P_i=P_{ii}\}$ i.e. the a-prior state. Furthermore, significant computation is involved with keeping the connections of such a large factor graph up to date i.e. graph management (based on updating occupancy values). Therefore, it is pertinent to consider a local graph construction scheme that iteratively redefines the inference problem dependent on the expected amount of information gained by performing said inference. This gives rise to a chicken and egg problem, where we do not know the problem formulation before performing inference, something which is impossible with conventional direct solvers. However, the message passing of \ac{GaBP} does allow us to achieve this. The rest of this section will outline our third contribution, which is a graph management methodology for redefining a local factor graph whilst inference is being performed upon it.

To achieve iterative graph construction, we can take inspiration from iterative factor graph problems such as those generated in pose optimisation for \ac{SLAM}. In these works at each iterative step, new sensor measurements and landmark observations are added sequentially to the factor graph and connected to a new corresponding state estimate and/or a previous state estimate in the case of loop closure. Whilst this process can be mirrored in \ac{GDM} by adding a new graph node for each new sensor reading at a previously unvisited location, it is not immediately obvious how far this sensor measurement will effect is local surroundings. It would be feasible to heuristically add a graph support to the initial voxel location, but this would not take into account the current belief and therefore the information actually added by the said measurement. We can however take advantage of the properties of wildfire propagation to expand our factor graph in an information theoretic manner, as messages propagate out from the measurement. The following outlines the process of this graph expansion.

When a new measurement $z_{k,i}$ is recorded from a sensor, whose corresponding node $x_i \notin \mathcal{X}$, a new variable node is added to the factor graph i.e. $\mathcal{X} \xleftarrow{+} x_i$, registered at the given sensor location. The six possible neighbours of $x_i$ are then added to the neighbourhood of $x_i$ given there is no obstacle blocking the connection between any pair of voxels. The information matrix $\mathbf{\Lambda}$ and observation vector $\mathbf{g}$ are updated using Eq. \eqref{eqn:info_ii} - \eqref{eqn:g_i} to account for the new nodes in the factor graph. The wildfire iteration can then occur with some modification to allow for dynamically expanding the factor graph around the new information. As per the standard wildfire process, when a message is sent to neighbour $j$ from node $i$, if the residual is greater than the wildfire threshold $r_m(m_{ij}) > \epsilon$, the neighbour $j$ is added to the queue $Q\xleftarrow{+}j$ for subsequent transmission (Alg. \ref{alg:sched}, Lines 8-10). However, at this point the neighbour $j$ may not have been expanded i.e., $\mathcal{N}_j = i$, and therefore cannot propagate any new information further. In this case, the node $j$ should then be checked for potential expandable neighbour nodes, $e$, to add to the factor graph $\mathcal{X} \xleftarrow{+} x_e$, $\forall \, e \in \mathcal{N}_j \land x_e \notin \mathcal{X}$. This process continues as the wildfire propagates and the factor graph grows iteratively until the outer nodes of the factor graph are receiving less information than the wildfire threshold $\epsilon$. This then couples the size the graph to the information theoretic value of a given measurement. Due to the fact that both information propagation and graph construction depend on the value of $\epsilon$, selection of this value is important for efficient convergence and adequate graph coverage. 

Fig. \ref{fig:graphGrowth} shows an example 2D iterative growth process pictorially. In this scenario the measurement $z_k$ is added to a discrete point in 2D space and an associated node $x_i$ is generated along with 3 neighbours $\{x_{j_1},x_{j_2},x_{j_3}\}$. Note that a possible neighbour to the left has not been added due to the presence of an obstacle between $x_i$ and the potential $x_{j_4}$ location. The information from $x_i$ is then sent to each of the neighbours, wherein both $x_{j_1}$ and $x_{j_2}$ receive information above threshold $\epsilon$. These nodes are then added to the queue $Q\xleftarrow{+}\{x_{j_1},x_{j_2}\}$, and then their respective neighbours $x_e \in \mathcal{N}_j$ are added to the graph (green nodes) if $x_e \notin \mathcal{X}$. Conversely, $x_{j_3}$ does not receive above the threshold information gain, and therefore its potential neighbours are not added to the graph. Also note, that a independent part of the factor graph (purple nodes) from a previous measurement ($z_{k-1}$) may also be connected to the new expansion and will iterate messages as per Alg. \ref{alg:sched}. \textcolor{black}{If the message information does not propagate enough to connect to the main graph, then the independent sub-graph may also remain disconnected forming two independent graphs. Due to message passing only needing neighbour information, having disconnected graphs makes no difference to the global solution and can be reconnected at any time.} 

In the circumstance where a new measurement does have an associated node $z_{k,i}$, then belief propagation occurs as normal until an unexpanded node is met, after which its neighbours are attempted to be added as per above.

\subsection{Proposed algorithm}

\begin{algorithm}[htb]
    \label{alg:proposed}
    \SetAlgoLined
    \caption{Proposed G-\ac{GaBP} solver}
    Define factors $f_p, f_o, f_d$ and $z_0 = 0$ \\
    Initialise empty factor graph, $\mathcal{X}, \mathbf{\Lambda}, \mathbf{g} \gets \emptyset$ \\
    \While{\text{Active}}{
        \If{New $z_{k,i}$}{
            \tcp{Update factor graph}
            \If{$x_i \notin \mathcal{X}$}{
            $\mathcal{X} \xleftarrow{+} x_i$ \\
            Define $\mathcal{N}_i$ based on $o_{ij}$\\
            $\mathcal{X} \xleftarrow{+} x_j,$ $\forall j \in \mathcal{N}_i$ \\
            Add new edge potentials: $P_{ij}=0$, $\mu_{ij}=0$, $m_{ij}^{k-1}=\{1/\sigma^2_p,0\}$\\
            }
            Update $\mathbf{\Lambda}$ and $\mathbf{g}$ using \eqref{eqn:info_ii}-\eqref{eqn:g_i} \\
            Update self potentials $\phi_i$: $P_{ii} = \mathbf{\Lambda}_{ii}$, $\mu_{ii}=\mathbf{g}_i/\mathbf{\Lambda_{ii}}$\\
            \tcp{Wildfire iterations}
            \For{$i \gets z_{k,i}$ : $z_{1,i}$}{
                $Q \xleftarrow{+} i$ \\ 
                \While{$Q \neq \emptyset$}{
                    $t \gets Q(1)$ \\
                    Update and send $m_{tj}$ according to \\
                    Eq. \eqref{eqn:messageUpdate}-\eqref{eqn:messageSend} $\forall j \in \mathcal{N}_t$\\
                    \If{$r_m(m_{ij}) > \epsilon$, \eqref{eqn:resid}}{
                        \If{$x_j$ \text{not expanded}}{
                            \tcp{Online graph expansion}
                            Define $\mathcal{N}_j$ based on $o_{je}$\\
                            $\mathcal{X} \xleftarrow{+} x_e$, $\forall$ $e \in \mathcal{N}_j \land x_e \notin \mathcal{X}$ \\
                            update $\mathbf{\Lambda}$ and $\mathbf{g}$ \\
                            Update self potentials: $P_{ii}$, $\mu_{ii}$ \\ 
                            Add new edge potentials: $P_{je}=0$, $\mu_{je}=0$, $m_{je}^{k-1}=\{1/\sigma^2_p,0\}$\\
                        }
                        \If{$j \notin Q$}{
                            $Q \xleftarrow{+} j$
                        }
                    }
                    $Q \xleftarrow{-} t$ \\
                }
            }
        }
        \tcp{Residual propagation}
        $m_{tj} = \argmax_m r_m(\mathbf{m})$ \\
        Update and send $m_{tj}$ based on Eq. \eqref{eqn:messageUpdate}-\eqref{eqn:messageSend} \\
    } 
    
\end{algorithm}

\begin{figure*}[t]
    \centering
    \includegraphics[width=0.8\textwidth]{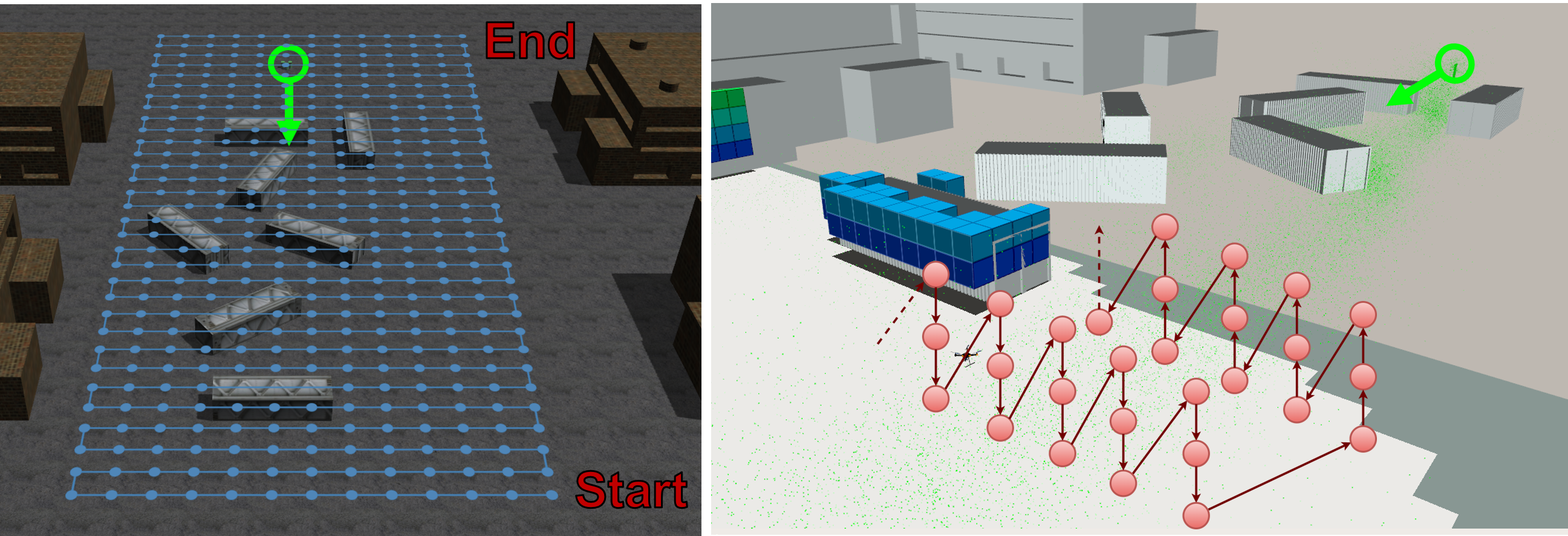}
    \caption{Simulation scenario. 2D sweep is shown left with black trajectory and fixed sampling waypoints (circles). The green circle and arrow shows the source location and wind direction respectively. Pictured right is the simulated UAV operating with a representative 3D sawtooth sweep (red circles and trajectory) for the 3D mapping task. Also shown is the online Octomap being built as well as instantaneous gas particles from GADEN (green dots).}
    \label{fig:sim_scenario}
\end{figure*}

As the proposed methodology shifting away from an a-priori graph definition towards a dynamic inference problem, we now introduce the proposed G-\ac{GaBP} algorithm, whose pseudo-code is provided in Alg. \ref{alg:proposed}.

Firstly, hyper-parameters of the problem are set a-priori and the empty factor graph is initialised. When a new measurement is received, if $z_{k,i}$ has no corresponding $x_i \in \mathcal{X}$, then a new node is added along with its immediate neighbours (lines 6-8). When adding these connections to the factor graph, new edge potentials are initialised which define the incoming information between the new nodes (line 9). During this step, the prior message information, $m^{k-1}_{ij}$, must be set to some minimum amount of information that can be received along an edge. This is defined as $\mathcal{N}(0,\sigma^2_p)$ by studying Eq. \eqref{eqn:messageSend} and the case of a factor graph consisting of a chain of nodes at an infinite distance from an observation. This is required since if this value is initially set to \{0,0\}, then calculating Eq. \eqref{eqn:resid} results in infinite information and therefore the propagation of growing the factor graph will occur indefinitely until the entire area is represented in the factor graph (nullifying the local graph growing procedure).

The information matrix $\mathbf{\Lambda}$ and the observation vector $\mathbf{g}$ are then updated to reflect any new nodes added to the factor graph and also to recalculate the time dependent precisions of Eq. \ref{eqn:info_ii} and time dependent observations of \eqref{eqn:g_i} (line 11). The self potentials required for belief propagation are then also updated to reflect this change (line 12). 

Line 13 then starts the wildfire propagation process. The node associated with the latest measurement i.e., $z_{k,i}$, is first added to the queue for propagation. Propagation then occurs according to the wildfire schedule, with the factor graph iteratively expanding with the propagating information (lines 15-32). If the time dependent decay factor has been employed, after the first measurement has propagated its information, then all historic measurements must also trigger their own wildfire iterations to transmit the now reduced information which they carry. Since residuals have already converged around previous measurements, these subsequent wildfire iterations usually require very little message passing.

Once all measurements have converged, residual belief propagation is performed until a new measurement is received (lines 35-36). If a new measurement is received before this stage, the process can be stopped at any point and the new measurement inserted (the anytime property of \ac{GaBP}).

This concludes the algorithmic portion of the paper. The proposed solver of Alg. \ref{alg:proposed} has also been developed in python as an open source ROS package \footnote{\ac{GaBP} open source ROS  package available from \url{https://github.com/callum-rhodes/gabp\_mapping}} for implementation on real and simulated mobile robots (for both 2D and 3D inference). 

\section{Simulation study} \label{sec:sim}

To showcase the real-time application of the 3D-GDM solver for use with a mobile robot, we deploy the algorithm in an example online mapping scenario. The setting for the simulation is a time-varying source release in a $200\times100\times20$m industrial arena consisting of buildings and shipping containers. The ground truth concentration data is modelled using GADEN \cite{Monroy2017}, a 3D filament dispersion simulator capable of modelling gas dispersions in feature rich environments. The specific release for the scenario is an acetone source situated in \ac{CFD} derived wind flow field with release rate of 1000ppm/s. Whilst the source release is modelled as an unsteady process, due to the constant wind field, the main structures of the plume remain consistent and therefore the mapping process is approximately time invariant. The sensor response behaviour is modelled inside GADEN and is inherently noisy (see \cite{Monroy2017} for further details on the simulator). The simulated mobile sensing agent is a quadrotor with a \ac{PID} affixed to the base. Realistic physics and flight control are handled using the popular simulation environment Gazebo and the PX4 autopilot plugin respectively. The operating scenario of the mapping task is given in Fig. \ref{fig:sim_scenario}.


\subsection{2D simulation study}
\begin{figure*}[t]
    \centering
    \includegraphics[width=0.75\textwidth]{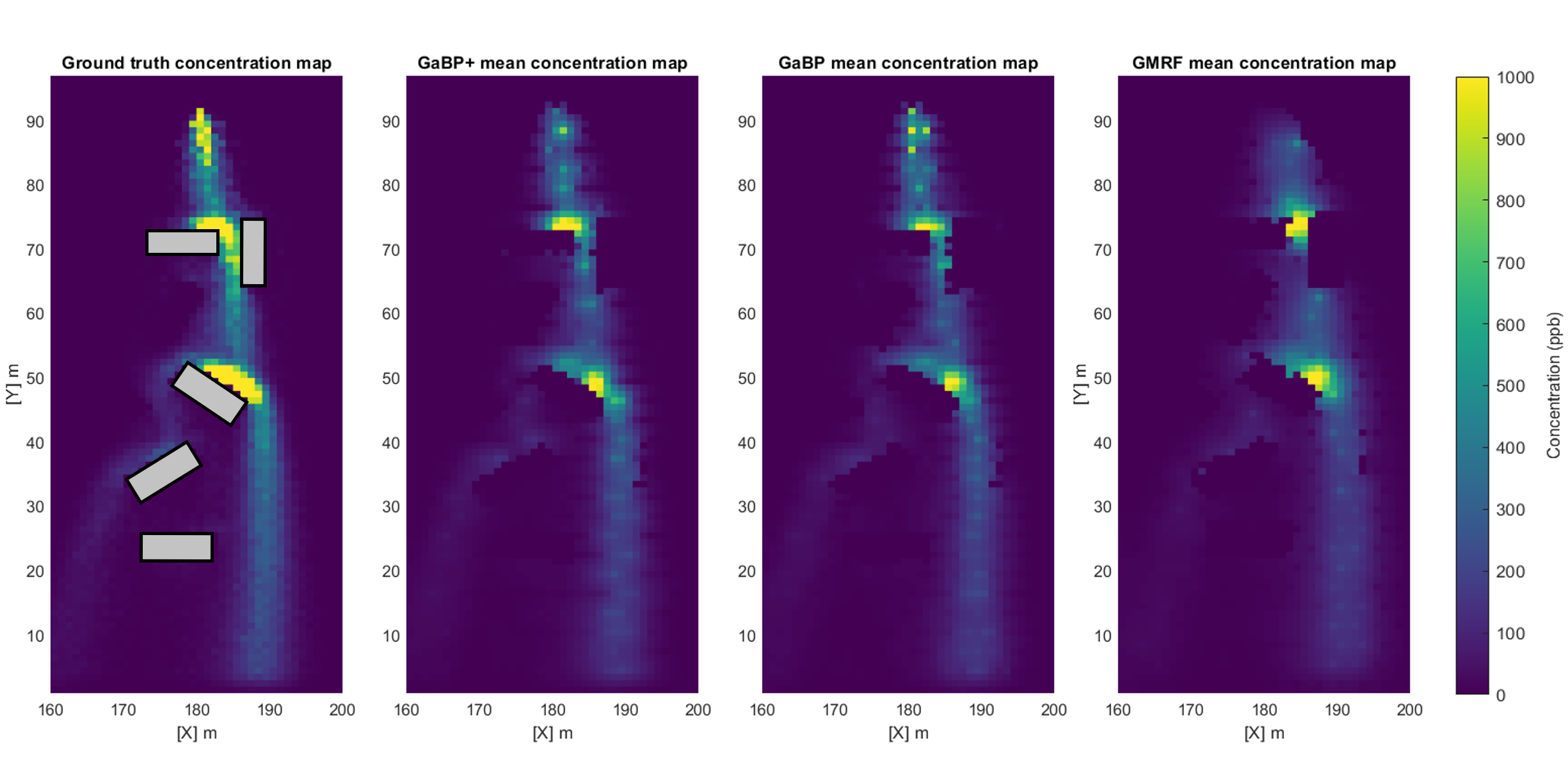}
    \caption{Final marginal mean estimates of the plume region for each algorithm compared against ground truth values. \textcolor{black}{Obstacle locations shown as grey rectangles.}}
    \label{fig:sim_2D_mean}
\end{figure*}

To benchmark the algorithm against the direct methods, we first deploy the algorithm in a 2D inference task over the entire area. This is necessary due to the scale of the area making direct solvers incapable of performing 3D inference. For example, at 1m resolution the size of the corresponding \ac{GMRF} for the 3D simulation scenario is a $4\times10^5$-by-$4\times10^5$ information matrix, which takes approximately 6 minutes to resolve using a normal PC. This is clearly unsuitable for online mapping tasks. We also make comparison to our original G-\ac{GaBP} algorithm \cite{Rhodes2022b}, without the optimisations proposed for the 3D-GDM solver. To differentiate between those two algorithms, we dub the proposed solver as \ac{GaBP}$+$ and the original as \ac{GaBP}.

For data collection, the \ac{UAV} flies at 1m/s along a pre-determined sweeping trajectory through the plume area with fixed sampling waypoints that must be sampled (at 3m separation distance). If the \ac{UAV} encounters an obstacle, it performs obstacle avoidance manoeuvres and any measurements taken above the estimation plane whilst avoiding obstacles are not included. The concentration sensor publishes at a constant rate of 2Hz and the mapping algorithm can include measurements that are captured when travelling between the waypoints. It is also enforced that at a minimum, the \ac{UAV} must resolve the measurement at each fixed sampling point into the distribution map, before sampling the next location. This constraint is chosen to represent the scenario where a decision about the next course of action may be taken based on the updated belief of the dispersion (as would be required by informative path planning or dictated by a human operator). It also allows the dataset $\mathcal{Z}$ to cover the search area evenly. This arrangement means that once a sample is taken at a waypoint, the measurement is inserted into the solver and whilst resolving said measurement, the \ac{UAV} will start to move to the next waypoint. If by the time the \ac{UAV} reaches the next waypoint, that the previous sample has not been resolved into the map, the \ac{UAV} will wait before sampling the current location. Conversely, if the solver is capable of resolving the first measurement before reaching the next waypoint, then any intermediary measurements taken en-route may be included in the inference as soon as the solver becomes available to accept a new measurement, i.e., sensor measurements must be sequential. For \ac{GMRF} this is defined when the \ac{MAP} solution of $p(\mathbf{x}|\mathcal{Z})$ is retrieved, and for \ac{GaBP} this is defined as when the wildfire iteration for all current $z_{k,i} \in \mathcal{Z}$ completes. The ability to include intermediary measurements is chosen to demonstrate the effective online performance of the \ac{GaBP} based solvers. The ground truth data is also gathered for error comparison, which records the average concentration over the simulation at each 1m grid cell over time. For context, if a single mobile sensor was to collect the ground truth data with a 10s sampling period, the data collection would take over 12 hours for the 2D ground truth (and over 60 hours for 3D), whereas the expected search time for the 2D sweep is approximately 1700s (within typical flight budgets for quadrotor UAVs). From an application point of view, this shows the necessity for such dense mapping algorithms to exist for hazardous material first response.

\begin{figure}[!ht]
    \centering
    \includegraphics[width=0.75\columnwidth]{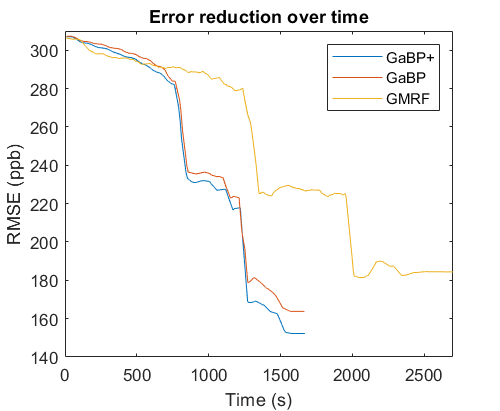}
    \caption{2D simulation RMSE reduction over time for each of the 3 tested algorithms}
    \label{fig:sim_2D_rmse}
\end{figure}

Hyper-parameters for the \ac{GMRF} model are kept consistent between the 3 algorithms with $\sigma^2_r=2$, $\sigma^2_s=0.1$, $\sigma^2_d=1e^{4}$ and for the \ac{GaBP} solvers, $\epsilon=0.01$. The time varying factor is set as $\sigma^2_\zeta=\infty$ since we are modelling a quasi-steady state plume structure and therefore measurements are time independent (but still subject to noise). Fig. \ref{fig:sim_2D_mean} shows the resultant marginal mean concentration maps after completing the search, and Fig. \ref{fig:sim_2D_rmse} the overall reduction in \ac{RMSE} over time, for each of the 3 algorithms. Numerical statistics of the mapping task for each algorithm are shown in Tab. \ref{tab:sim_stats}. \textcolor{black}{Metrics for comparing uncertainty outputs between \ac{GMRF} and \ac{GaBP} are not included since their output values are not comparable, as shown in \cite{Rhodes2022b}.}

Only the region of the interest (the plume) is used for the \ac{RMSE} calculation, defined as the set of cells, $\mathcal{X}_{\text{plume}}$, exhibiting higher than a background level concentration, $z_{thresh} = 100\text{ppb}$, i.e., $\mathcal{X}_{plume} \gets x_i \in \mathcal{X}$ $|$ $\mu_{gt}(x_i) > z_{thresh}$ where $\mu_{gt}(x_i)$ is the time averaged ground truth concentration at the location associated with variable $x_i$. This is so that the noise associated with modelling the zero concentration background regions does not dominate the output \ac{RMSE} of each algorithm. The calculation for \ac{RMSE} is therefore:
\begin{equation}
\text{RMSE}=\sqrt{\frac{\sum_{i=1}^{n_{plume}}(\mu_{gt}(x_i)-\mathbb{E}(x_i))^2}{n_{plume}}}
\label{eqn:rmse}
\end{equation} 
where $n_{plume}$ refers to the size of the set $\mathcal{X}_{plume}$.

\begin{table*}
    \centering
    \caption{Mapping statistics for each algorithm for both the 2D and 3D simulations}
    \label{tab:sim_stats}
    \begin{tabular}{l|ccccc}
        \hline
        Algorithm   & Total         & Avg. resolve time     & Processed                         & Final (avg) number of             & Converged\\
        \quad       & runtime       & per measurement       & measurements, $|\mathcal{Z}|$      & estimated states, $|\mathcal{X}|$  & RMSE (ppb)\\
        \hline
        \hline
        2D-GMRF        & 2659s         & 5657ms             & 470                               &  25228 (25228)                    & 184 \\
        2D-GaBP        & 1644s         & 18ms              & 3253                              &  25228 (25228)                    & 163\\
        2D-GaBP+       & 1644s         & 14ms              & 3253                              &  6536 (4466)                      & 152\\
        \hline
        \hline
        3D-GaBP        & 4411s         & 15ms                 & 8834                           &  108914 (108914)                  & 197\\
        3D-GaBP+       & 4400s         & 13ms                 & 8998                           &  46292 (28058)                    & 186\\
        \hline
    \end{tabular}
\end{table*}

The resultant marginal mean concentration maps show very similar structures of the plume to the ground truth, with variations in peak values (particularly near the source) due to the chaotic nature of gas dispersion and a short sampling time. Whilst the mean maps show similar features, when appreciating the \ac{RMSE} relative to the ground truth it can be seen that G-\ac{GaBP}+ has the best final state estimate and error reduction over time due to the hybrid solver taking advantage of downtime between wildfire iterations. Because of this, we also see that measurements are also resolved slightly quicker (14ms vs 18ms), since they are being inserted into a more converged belief. This improved performance is managed whilst only estimating one quarter the number of states (6536 vs 25228), significantly reducing the computational overhead in managing the factor graph structure, which is linearly related to the number of estimated states. The direct \ac{GMRF} solver in contrast to the \ac{GaBP} solvers has much worse online performance. In our simulation, this is directly caused by the slow update rate of the \ac{GMRF} solver and therefore significantly fewer measurements are recorded in the final map (460 vs 3253). Whilst \ac{GMRF} takes an average of 4.8s to resolve the map per measurement, \ac{GaBP}+ is capable of solving a single measurement on average in 14ms, significantly quicker than the update rate of the sensor and hence why \ac{GaBP} and \ac{GaBP}+ process the same number of measurements (i.e. all of the measurements output by the sensor). 

Reaffirming our conclusions in \cite{Rhodes2022b}, \ac{GaBP} propagation has shown its superior performance for online inference against direct solvers. The following simulations explore the online obstacle aware 3D mapping abilities of the proposed algorithm, something not yet achieved in the literature.

\subsection{3D simulation study}

\begin{figure*}
    \centering
    \includegraphics[width=0.7\textwidth]{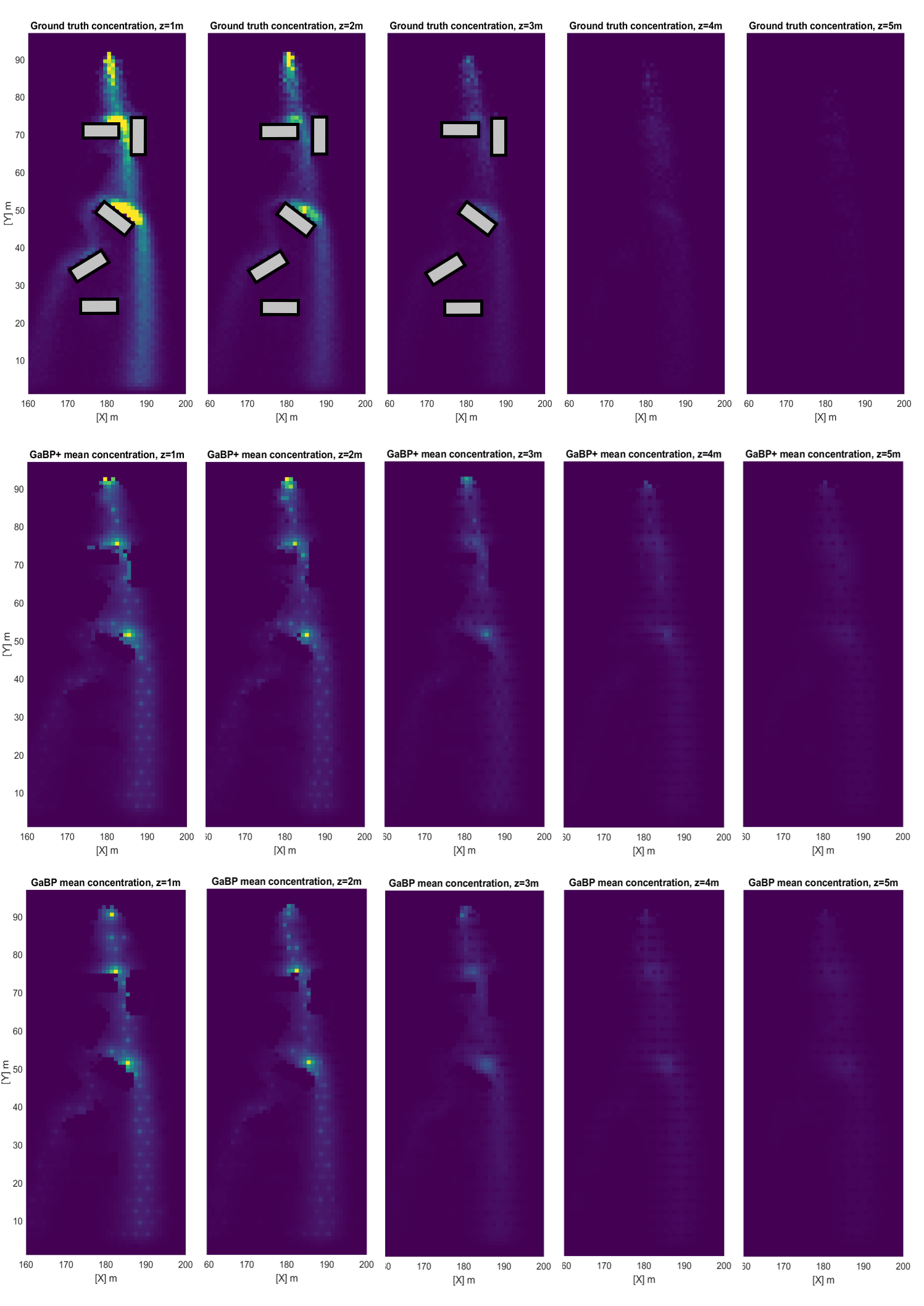}
    \caption{Final marginal mean estimates of the search for 3D-GaBP \& 3D-GaBP+ compared against ground truth values at $z=1,2,3,4,5$m. \textcolor{black}{Obstacle locations show as grey rectangles at heights where they are present.}}
    \label{fig:sim_3d_mean}
\end{figure*}

For the 3D study, we explore the same scenario however expand the state estimation to include the 3D advancements of section \ref{sec:GDMfactors}. This expands the possible estimation set $\mathcal{X}$, to be up to $4\times10^5$ nodes which are also more densely connected than the 2D case. As explained previously, \ac{GMRF} is unsuitable for online state estimation in this scenario and therefore \ac{GaBP} and \ac{GaBP}+ are compared together. The graph management overhead for the original \ac{GaBP} algorithm is also computationally limited when considering the full $200\times100\times20$m scenario area and therefore only the area covered by the sweep is considered in the 3D task (furthermore showing the necessity of the contribution in section \ref{sec:local}). Parameters of the search from the 2D study are carried forward however a saw-tooth sweep in the vertical axis z$=\{1,3,5\}$m is performed to collect measurements required for 3D mapping (see Fig. \ref{fig:sim_scenario}). This extends the expected search time to 70 minutes.

\begin{figure}[H]
    \centering
    \includegraphics[width=0.75\columnwidth]{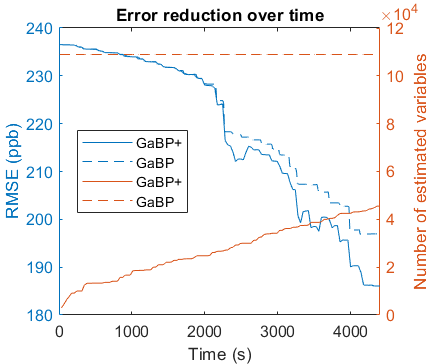}
    \caption{3D simulation RMSE reduction over time for GaBP+ and GaBP. Also shown is the number of estimated variables, $|\mathcal{X}|$, over time.}
    \label{fig:sim_3D_rmse}
\end{figure}

The mapping results are given in Fig. \ref{fig:sim_3d_mean}, which shows 1m slices of the estimated 3D map compared against the ground truth data. Fig. \ref{fig:sim_3D_rmse} shows the \ac{RMSE} reduction over time. Other Statistical data are provided in Tab. \ref{tab:sim_stats}.

As with the 2D task, concentration estimates across the map are visually comparable with the ground truth and the difference between \ac{GaBP} and \ac{GaBP}+ is minimal. However, when viewing the \ac{RMSE} reduction rate it is clearly shown that the proposed 3D-\ac{GDM} algorithm shows a numerically greater estimation accuracy over time, and does so with far fewer number of estimated states. As with the 2D case, the average resolve time per measurement is very low for both algorithms (less than the 2D case) with \ac{GaBP}+ being marginally faster than the original as previously discussed. This is because the plume above 3m is not very different from the a-prior state ($\mu = 0$) and therefore measurements taken above 3m are close to 0 and do not take long to converge (reducing the average measurement resolve time). This is somewhat counter-intuitive as it would be expected that a 3D mapping task would be more complicated and take longer to converge. However, wildfire propagation is scale agnostic and its information theoretic nature allows areas of low information to not negatively affect the solve time. Furthermore, the impact of obstacles on the 3D gas distribution is fully modelled as zero concentration readings are predicted in the obstacle locations at heights of z$=\{1,2,3\}$ but then smooth predictions are shown in these same areas for variables at height z$=\{4,5\}$.  

This concludes the simulation study and it has been clearly demonstrated the advantage of the \ac{GaBP} over conventional direct solvers. In addition, the locally expanding inference and hybrid schedule proposed in this paper have demonstrated a significant improvement on estimate convergence as well as on the computational efficiency of graph management. These additions have decoupled the problem definition from a predefined search area allowing for iterative, fast and local 3D distribution mapping, a major step forward for complex online \ac{GDM} for mobile sensing.

\section{Experimental trial} \label{sec:experiment}

\begin{figure*}
    \centering
    \includegraphics[width=0.7\textwidth]{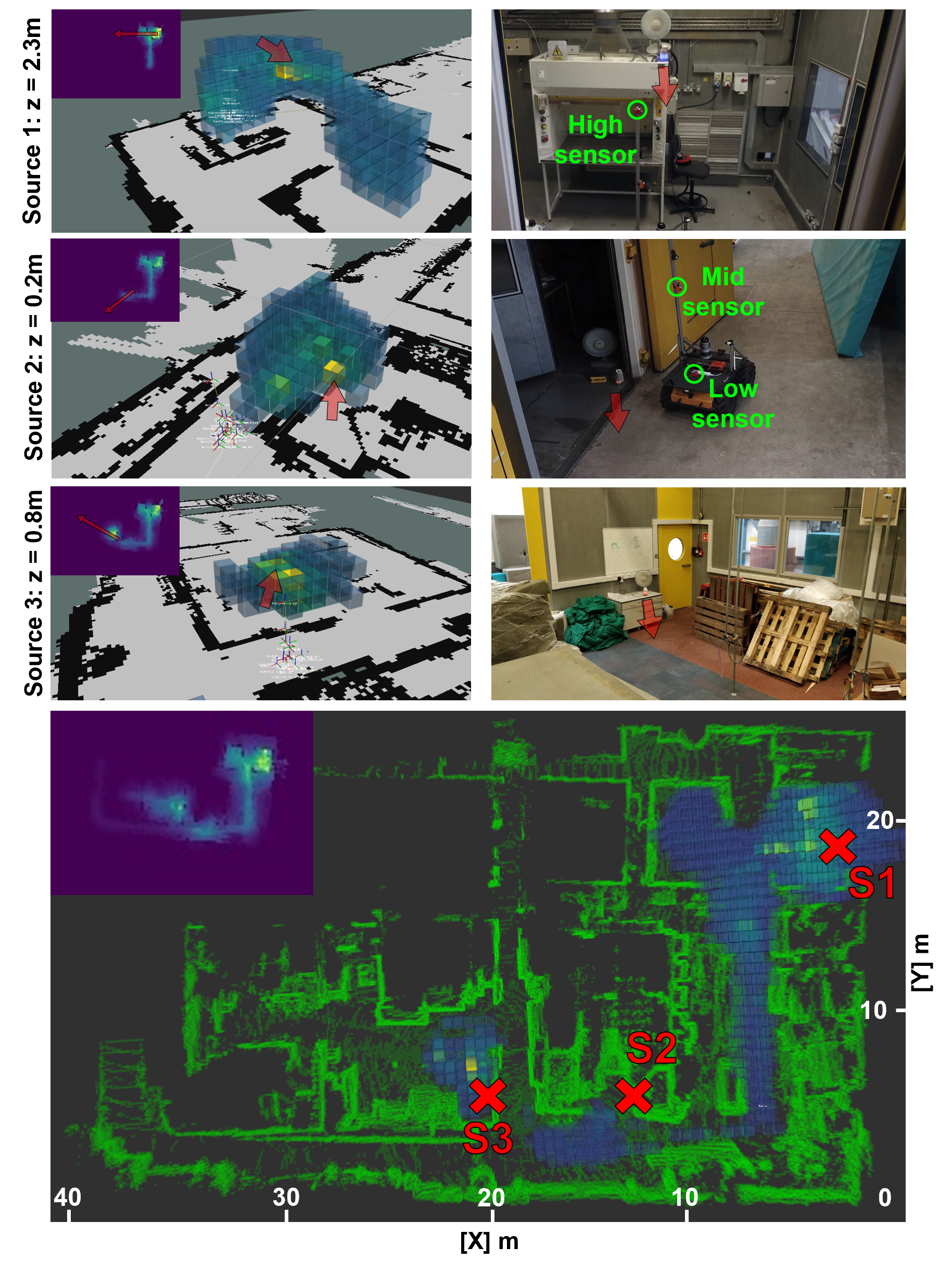}
    \caption{Experimental application of the 3D G-GaBP solver in a cluttered GPS-denied scenario. Each estimated 3D plume structure for the 3 sources are shown \textcolor{black}{(coloured voxels for any cell with a concentration above 20\% of the maximum concentration value, yellow indicates high concentration)}. The corresponding 2.5D map at the point where the source is driven past is shown in the top left corner. The base of each red arrow indicates the source location and the direction represents the fan induced wind direction. The real-world location of each source plume is shown on the right for context. Shown bottom is the full 3D \& 2.5D distribution maps superimposed on the Octomap.}
    \label{fig:GW3}
\end{figure*}

Having shown the benefit of the proposed 3D-GDM solver against the current state of the art in simulation, we now apply the algorithm in a real-world 3D mapping task to show its capability. One of the key benefits of using the \ac{GaBP} solver is the ability to generate real time mapping solutions onboard the local hardware of the sensing agent, whereas all work previously has been performed offboard. This section will demonstrate online 3D mapping entirely onboard a ground vehicle (with networking only used for monitoring), and is the first practical application of the \ac{GaBP} solver for \ac{GDM}.

The scenario for the task is a cluttered industrial workshop with several isolated work units within the area. Three source releases are placed within the environment at different heights (this is to discern the concentration levels along the z-axis). The source releases are beakers of liquid acetone, agitated by a ultrasonic bubbler and dispersed using a desk fan. 

To perform data collection, a Clearpath Husky \ac{UGV} is fitted with a sensing pole that has \ac{PID} sensors affixed at 0.3, 1 and 1.8m. Sensor data is recorded at 1Hz for each sensor and fed to the \ac{GaBP} solver in an adhoc manner. The \ac{UGV} is manually teleoperated around the environment in an exploratory search with live feedback of the marginal mean and uncertainty maps. Online 3D obstacle maps are built using the Lidar outputs of a Velodyne VLP-32C and a Robosense-Bpearl to achieve nearly full $360^o$ coverage of the surroundings. The 3D map is generated using the popular 3D map representation and ROS package Octomap \cite{Hornung2013}, and it is occupancy values from this software that directly condition the factor graph structure defined in Eq. \eqref{eqn:info_ii} - \eqref{eqn:g_i}. Localisation of the robot is achieved using the Velodyne VLP-32C in conjunction with the open source graph \ac{SLAM} package hdl-graph-slam \cite{Koide2019}. Good localisation is fundamental for accurate GDM mapping, not only because pointclouds are inserted into the Octomap relative to the local pose (i.e. mapping with known poses), but also the pose estimate informs the measurement factor association with a specific state variable (i.e. defines $z_{k,i}$). The 3D G-GaBP algorithm is run with the same hyper-parameters as in the simulations however with a higher resolution of 0.5m and a higher $\epsilon = 1$. The approximate size of the area is $40\times25\times10$m.

Contrary to many other \ac{GDM} works, the proposed experimental platform does not rely on a external positioning system or an offboard processor, and therefore the platform can be deployed in any scenario in complete isolation. To the best of the authors' knowledge, this is the first time such a versatile platform has been deployed for the task of \ac{GDM}, aside from the academic contributions of the proposed algorithm. For a video of the 3D-GDM experimental trial, please visit \url{https://youtu.be/crAJd4afW8c}.

Fig. \ref{fig:GW3}, shows the 3D plume predicted by the proposed algorithm for each one of the 3 sources as the robot is driven past each source. The 3D aspect of each sources plume can be easily discerned when studying the coloured voxel concentration map. Source 1 is the first source encountered and is situated above all three sensors at 2.3m. A high concentration is measured next to the source location (bright yellow voxel) but nearly zero chemical is detected on the bottom sensor at the same location, this results in a plume that is dispersed above the floor settling in the opposite room at floor level. We can also see significant mixing behind the source at the back of the room. Source 2 is encountered next, and this source is dispersed along floor level at 0.2m. Only the low sensor reads a high concentration value and therefore the distribution shows the plume spreading along the floor, with some mixing at 1m. To reinforce the importance of the 3D modelling aspect, a 2.5D map is also shown in the top left corner at the point in time the source is sampled. To generate the 2.5D map, concentration values from the 3D inference are averaged along the z-axis and projected down to their corresponding x-y coordinate. If using this map alone to analyse the gas distribution, then the high altitude source dispersion is clear, however the low height source is masked by the low concentration readings of the mid and high sensors and therefore cannot be discerned as a possible high concentration area (whereas this source is clear using the 3D plume map). This is a significant advantage of using a 3D inference model over a 2D model since source 2 may have been overlooked if using the 2D map alone. Finally, source 3 is sampled in a larger room and the plume is shown to disperse up to about 1.5m high. The low concentration readings of the high sensor stop the distribution propagating upwards where the acetone does not mix (due to a higher density than air). The complete distribution map is shown at the bottom of Fig. \ref{fig:GW3} after the robot has searched the area. Due to the density of vapourised acetone and its propensity to settle on surfaces, we can see that the low and mid sources do not show as much dispersion as the high source, which is able to be transported much further through the environment. From this image we can also see the requirement for obstacle aware \ac{GDM} in such a cluttered environment, since for both source 1 and source 3 there is no mixing of the high concentration to the other side of their surrounding walls (and equally the low concentration readings outside the rooms do not directly interact with the high concentration plume inside the rooms). 

Mapping statistics for the entire experimental mapping task are shown in Tab. \ref{tab:exp_stats}. Due to the fact that several returning passes of the rooms are made during the 15 minute search (and a higher $\epsilon$ value), the average resolve time per measurement is significantly lower than the simulation tests since the main plume structures are established within the first pass, and then only slightly updated and refined in subsequent traversals.

\begin{table}
    \centering
    \caption{Mapping statistics from the experimental trial}
    \label{tab:exp_stats}
    \begin{tabular}{cccc}
    \hline
    Runtime         & Avg. resolve time     & $|\mathcal{Z}|$      & $|\mathcal{X}|$  (average)      \\
        \hline
        \hline
        942s         & 3ms             & 5467              &  18534 (13826)\\
        \hline
    \end{tabular}

\end{table}


\textcolor{black}{\section{Future works} 
The work encompassed in this paper has demonstrated the real-world applicability of the belief propagation solver to an environmental monitoring task with a mobile sensor. Having validated that the solver can perform large scale, real time and obstacle aware modelling onboard a standalone platform, efforts can now be made to further develop the system and make use of the computational advantages brought forward by \ac{GaBP}.
One of the key areas this work can be extended is in more complex modelling tasks. A natural extension is to included wind measurements alongside concentration measurements, and to model both phenomena and their mutual influence concurrently, such as that proposed by Gongora et. al \cite{Gongora2020}. It is foreseen that belief propagation will excel in this case, as \ac{GaBP} has already shown to provide excellent solve times for other non-linear optimisation tasks, as shown by Ortiz et. al \cite{Ortiz2021}.
Furthermore, predictive inference for \ac{IPP} using the \ac{GaBP} framework now becomes computationally achievable, and with the ability to have a fixed time cost. However, due to the the convergence of estimates also factoring into the quantified uncertainty output, how to apply information theoretic criteria remains an open question.
Finally, distributed systems may be explored wherein a multi-robot system uses the \ac{GaBP} framework to update neighbouring robots of shared beliefs in an efficient manner. }

\section{Conclusions}

This paper first outlines the problem of performing inference for 3D gas distribution maps upon a connected factor graph. Due to the computational complexity of such a large problem, the \ac{GaBP} algorithm is applied as the iterative solver to achieve real time distribution maps as concentration readings are collected. The G-\ac{GaBP} algorithm is then extended over the original of \cite{Rhodes2022b} in two key areas. First, a hybrid message schedule is proposed that leverages the fast local convergence of the wildfire algorithm with the globally converging properties of residual belief propagation. Second, the \ac{GDM} problem is formulated incrementally as inference is performed, tightly coupling information propagation and graph construction in order to achieve entirely local inference. This feature also allows the proposed algorithm to be more easily integrated with popular SLAM solutions. 

These new contributions are then compared in simulation against the standard direct solver of G-\ac{GMRF} and the original G-\ac{GaBP} in both 2D and 3D mapping tasks. The proposed solver is found to be able to perform more accurate inference (due to the hybrid scheduler) whilst simultaneously estimating significantly less states (reducing computational burden on factor graph management). These enhancements allow the 3D G-\ac{GaBP} algorithm to be deployed onboard cheaper hardware units without the need for GPU acceleration. 

To prove this and show its real-worlds applicability, the algorithm is deployed onboard a \ac{UGV} with 3 gas sensors at varying heights. The vehicle is driven around a cluttered GPS-denied environment and in real time resolves complex structures of the gas distribution for 3 different sources placed at different heights in the environment. It is also shown how using a 2D distribution map can obfuscate features of the plume that are only identifiable using 3D inference. The development of an online 3D-\ac{GDM} solution opens up further research into exploring informative path planning as well as multi-agent mapping for operation in cluttered environments where gas distributions occur in scenarios exhibiting verticality. Given the fact that target gases often have densities that are higher or lower than air, this is a very real prospect that was previously infeasible. 



\bibliographystyle{IEEEtran}
\bibliography{IEEEabrv,library}

\begin{thebibliography}{10}
\providecommand{\url}[1]{#1}
\csname url@samestyle\endcsname
\providecommand{\newblock}{\relax}
\providecommand{\bibinfo}[2]{#2}
\providecommand{\BIBentrySTDinterwordspacing}{\spaceskip=0pt\relax}
\providecommand{\BIBentryALTinterwordstretchfactor}{4}
\providecommand{\BIBentryALTinterwordspacing}{\spaceskip=\fontdimen2\font plus
\BIBentryALTinterwordstretchfactor\fontdimen3\font minus
  \fontdimen4\font\relax}
\providecommand{\BIBforeignlanguage}[2]{{%
\expandafter\ifx\csname l@#1\endcsname\relax
\typeout{** WARNING: IEEEtran.bst: No hyphenation pattern has been}%
\typeout{** loaded for the language `#1'. Using the pattern for}%
\typeout{** the default language instead.}%
\else
\language=\csname l@#1\endcsname
\fi
#2}}
\providecommand{\BIBdecl}{\relax}
\BIBdecl

\bibitem{Towler2012}
J.~Towler, B.~Krawiec, and K.~Kochersberger, ``{Terrain and Radiation Mapping
  in Post-Disaster Environments Using an Autonomous Helicopter},'' \emph{Remote
  Sensing}, vol.~4, no.~7, pp. 1995--2015, 2012.

\bibitem{Branford2011}
S.~Branford, O.~Coceal, T.~G. Thomas, and S.~E. Belcher, ``{Dispersion of a
  Point-Source Release of a Passive Scalar Through an Urban-Like Array for
  Different Wind Directions},'' \emph{Boundary-Layer Meteorology}, vol. 139,
  no.~3, pp. 367--394, 2011.

\bibitem{Lilienthal2003}
A.~Lilienthal and T.~Duckett, ``{Gas Source Localisation by Constructing
  Concentration Gridmaps with a Mobile Robot},'' in \emph{Proceedings of the
  European Conference on Mobile Robots (ECMR 2003)}, 2003.

\bibitem{Lilienthal2009}
A.~J. Lilienthal, M.~Reggente, M.~Trinca, J.~L. Blanco, and J.~Gonzalez, ``{A
  statistical approach to gas distribution modelling with mobile robots - The
  Kernel DM+V algorithm},'' \emph{2009 IEEE/RSJ International Conference on
  Intelligent Robots and Systems, IROS 2009}, pp. 570--576, 2009.

\bibitem{reggente2009}
M.~Reggente and A.~J. Lilienthal, ``{Using local wind information for gas
  distribution mapping in outdoor environments with a mobile robot},''
  \emph{Proceedings of IEEE Sensors}, pp. 1715--1720, 2009.

\bibitem{Plagemann2011}
S.~Asadi, C.~Plagemann, C.~Stachniss, M.~Reggente, and A.~J. Lilienthal,
  ``{Statistical Gas Distribution Modeling Using Kernel Methods},''
  \emph{Intelligent Systems for Machine Olfaction}, vol.~1, no. March, pp.
  153--179, 2011.

\bibitem{G.Monroy2016}
J.~{G. Monroy}, J.~L. Blanco, and J.~Gonzalez-Jimenez, ``{Time-variant gas
  distribution mapping with obstacle information},'' \emph{Autonomous Robots},
  vol.~40, no.~1, pp. 1--16, 2016.

\bibitem{Gongora2020}
A.~Gongora, J.~Monroy, and J.~Gonzalez-Jimenez, ``{Joint Estimation of Gas {\&}
  Wind Maps for Fast-Response Applications},'' \emph{Applied Mathematical
  Modelling}, vol.~87, pp. 655--674, 2020.

\bibitem{Stachniss2009}
C.~Stachniss, C.~Plagemann, and A.~J. Lilienthal, ``{Learning gas distribution
  models using sparse Gaussian process mixtures},'' \emph{Autonomous Robots},
  vol.~26, no. 2-3, pp. 187--202, 2009.

\bibitem{Hutchinson2019}
M.~Hutchinson, P.~Ladosz, C.~Liu, and W.~H. Chen, ``{Experimental assessment of
  plume mapping using point measurements from unmanned vehicles},'' in
  \emph{Proceedings - IEEE International Conference on Robotics and
  Automation}, vol. 2019-May, 2019, pp. 7720--7726.

\bibitem{Bennetts2014}
V.~{Hernandez Bennetts}, E.~Schaffernicht, V.~Pomareda, A.~J. Lilienthal,
  S.~Marco, and M.~Trincavelli, ``{Combining non selective gas sensors on a
  mobile robot for identification and mapping of multiple chemical
  compounds},'' \emph{Sensors (Basel, Switzerland)}, vol.~14, no.~9, pp.
  17\,331--17\,352, 2014.

\bibitem{Luo2015}
B.~Luo, Q.~H. Meng, J.~Y. Wang, B.~Sun, and Y.~Wang, ``{Three-dimensional gas
  distribution mapping with a micro-drone},'' \emph{Chinese Control Conference,
  CCC}, vol. 2015-Septe, pp. 6011--6015, 2015.

\bibitem{Visvanathan2020a}
R.~Visvanathan, K.~Kamarudin, S.~M. Mamduh, M.~Toyoura, A.~S. {Ali Yeon},
  A.~Zakaria, L.~M. Kamarudin, X.~Mao, and S.~A. {Abdul Shukor}, ``{Improved
  mobile robot based gas distribution mapping through propagated distance
  transform for structured indoor environment},'' \emph{Advanced Robotics},
  vol.~34, no.~10, pp. 637--647, 2020.

\bibitem{Choi2012}
J.~Choi, M.~Jadaliha, and Y.~Xu, ``{Efficient spatial prediction using gaussian
  markov random fields under uncertain localization},'' in \emph{Dynamic
  Systems and Control Conference}, vol. 45318, 2012, pp. 253----262.

\bibitem{Dellaert2017}
F.~Dellaert and M.~Kaess, ``{Factor Graphs for Robot Perception},''
  \emph{Foundations and Trends in Robotics}, vol.~6, no. 1-2, pp. 1--139, 2017.

\bibitem{Rhodes2022b}
C.~Rhodes, C.~Liu, and W.-h. Chen, ``{Scalable probabilistic gas distribution
  mapping using Gaussian belief propagation},'' in \emph{2022 IEEE/RSJ
  International Conference on Intelligent Robots and Systems (IROS)}.\hskip 1em
  plus 0.5em minus 0.4em\relax IEEE, 2022, p. [awaiting publication].

\bibitem{Murphy2013a}
\BIBentryALTinterwordspacing
K.~Murphy, Y.~Weiss, and M.~I. Jordan, ``{Loopy Belief Propagation for
  Approximate Inference: An Empirical Study},'' \emph{arXiv preprint
  arXiv:1301.6725}, 2013. [Online]. Available:
  \url{http://arxiv.org/abs/1301.6725}
\BIBentrySTDinterwordspacing

\bibitem{Ranganathan2007}
A.~Ranganathan, M.~Kaess, and F.~Dellaert, ``{Loopy SAM},'' \emph{IJCAI
  International Joint Conference on Artificial Intelligence}, pp. 2191--2196,
  2007.

\bibitem{Davison2019}
\BIBentryALTinterwordspacing
A.~J. Davison and J.~Ortiz, ``{FutureMapping 2: Gaussian Belief Propagation for
  Spatial AI},'' \emph{arXiv preprint arXiv:1910.14139 (2019)}, 2019. [Online].
  Available: \url{http://arxiv.org/abs/1910.14139}
\BIBentrySTDinterwordspacing

\bibitem{Ortiz2020}
J.~Ortiz, M.~Pupilli, S.~Leutenegger, and A.~J. Davison, ``{Bundle Adjustment
  on a Graph Processor},'' \emph{Proceedings of the IEEE Computer Society
  Conference on Computer Vision and Pattern Recognition}, pp. 2413--2422, 2020.

\bibitem{Pearl1988}
J.~Pearl, \emph{{Probabilistic reasoning in intelligent systems: networks of
  plausible inference}}.\hskip 1em plus 0.5em minus 0.4em\relax Morgan
  kaufmann, 1988.

\bibitem{Shental2008}
O.~Shental, P.~H. Siegel, J.~K. Wolf, D.~Bickson, and D.~Dolev, ``{Gaussian
  belief propagation solver for systems of linear equations},'' \emph{IEEE
  International Symposium on Information Theory - Proceedings}, pp. 1863--1867,
  2008.

\bibitem{Ortiz2021}
\BIBentryALTinterwordspacing
J.~Ortiz, T.~Evans, and A.~J. Davison, ``{A visual introduction to Gaussian
  Belief Propagation},'' \emph{arXiv preprint arXiv:2107.02308, 2021}, pp.
  1--20, 2021. [Online]. Available: \url{http://arxiv.org/abs/2107.02308}
\BIBentrySTDinterwordspacing

\bibitem{Weiss2000}
Y.~Weiss and W.~T. Freeman, ``{Correctness of belief propagation in Gaussian
  graphical models of arbitrary topology},'' \emph{Advances in Neural
  Information Processing Systems}, vol. 2200, pp. 673--679, 2000.

\bibitem{Elidan2006}
G.~Elidan, I.~McGraw, and D.~Koller, ``{Residual belief propagation: Informed
  scheduling for asynchronous message passing},'' \emph{Proceedings of the 22nd
  Conference on Uncertainty in Artificial Intelligence, UAI 2006}, pp.
  165--173, 2006.

\bibitem{Liu2012}
X.~Liu, S.~Cheng, H.~Liu, S.~Hu, D.~Zhang, and H.~Ning, ``{A survey on gas
  sensing technology},'' \emph{Sensors (Switzerland)}, vol.~12, no.~7, pp.
  9635--9665, 2012.

\bibitem{Hutchinson2019a}
M.~Hutchinson, C.~Liu, and W.~H. Chen, ``{Source term estimation of a hazardous
  airborne release using an unmanned aerial vehicle},'' \emph{Journal of Field
  Robotics}, vol.~36, no.~4, pp. 797--817, 2019.

\bibitem{Rhodes2022}
C.~Rhodes, C.~Liu, and W.~H. Chen, ``{Autonomous Source Term Estimation in
  Unknown Environments: From a Dual Control Concept to UAV Deployment},''
  \emph{IEEE Robotics and Automation Letters}, vol.~7, no.~2, pp. 2274--2281,
  2022.

\bibitem{Rhodes2020}
C.~Rhodes, C.~Liu, and W.-h. Chen, ``{Informative Path Planning for Gas
  Distribution Mapping in Cluttered Environments},'' in \emph{2020 IEEE/RSJ
  International Conference on Intelligent Robots and Systems (IROS)}, 2020, pp.
  6726--6732.

\bibitem{Monroy2017}
J.~Monroy, V.~Hernandez-Bennetts, H.~Fan, A.~Lilienthal, and
  J.~Gonzalez-Jimenez, ``{GADEN: A 3D gas dispersion simulator for mobile robot
  olfaction in realistic environments},'' \emph{Sensors (Switzerland)}, vol.~6,
  no.~2, p. 1479, 2017.

\bibitem{Hornung2013}
A.~Hornung, K.~M. Wurm, M.~Bennewitz, C.~Stachniss, and W.~Burgard, ``{OctoMap:
  An efficient probabilistic 3D mapping framework based on octrees},''
  \emph{Autonomous Robots}, vol.~34, no.~3, pp. 189--206, 2013.

\bibitem{Koide2019}
K.~Koide, J.~Miura, and E.~Menegatti, ``{A portable three-dimensional
  LIDAR-based system for long-term and wide-area people behavior
  measurement},'' \emph{International Journal of Advanced Robotic Systems},
  vol.~16, no.~2, pp. 1--16, 2019.

\end{thebibliography}

\end{document}